\documentclass[floatfix,rmp,twocolumn,twoside]{revtex4}
\usepackage{graphicx}

\bibpunct{[}{]}{,}{n}{}{,}

\usepackage[english]{babel}
\usepackage{verbatim}
\usepackage{subfigure}
\usepackage{varwidth}
\usepackage{graphicx}
\usepackage{alltt}

\makeatletter
\g@addto@macro\@verbatim\small
\makeatother

\begin{document}

\title{Filling-Based Techniques Applied to Object Projection Feature Estimation}
\author{Luis~Quesada\\
  Department of Computer Science and Artificial Intelligence, CITIC, University of Granada, \\
  Granada 18071, Spain \\
  \textit{lquesada@decsai.ugr.es} \\ [3mm]
  Alejandro J. Le\'on\\
  Department of Software Engineering, CITIC, University of Granada, \\
  Granada 18071, Spain \\
  \textit{aleon@ugr.es} 
  }

\begin{abstract}

3D motion tracking is a critical task in many computer vision applications.
Unsupervised markerless 3D motion tracking systems determine the most relevant object in the screen and then track it by continuously estimating its projection features (center and area) from the edge image and a point inside the relevant object projection (namely, inner point), until the tracking fails.
Existing object projection feature estimation techniques are based on ray-casting from the inner point. These techniques present three main drawbacks: when the inner point is surrounded by edges, rays may not reach other relevant areas; as a consequence of that issue, the estimated features may greatly vary depending on the position of the inner point relative to the object projection; and finally, increasing the number of rays being casted and the ray-casting iterations (which would make the results more accurate and stable) increases the processing time to the point the tracking cannot be performed on the fly.
In this paper, we analyze an intuitive filling-based object projection feature estimation technique that solves the aforementioned problems but is too sensitive to edge miscalculations.
Then, we propose a less computing-intensive modification to that technique that would not be affected by the existing techniques issues and would be no more sensitive to edge miscalculations than ray-casting-based techniques.

\end{abstract}

\maketitle

\section{Introduction}
\noindent

Optical motion tracking, simply called motion tracking in this paper, means continuously locating a moving object in a video sequence.
2D tracking aims at following the image projection of objects that move within a 3D space.
3D tracking aims at estimating all six degrees of freedom (DOFs) movements of an object relative to the camera: the three position DOFs and the three orientation DOFs \cite{Lepetit2005}.

A 3D motion tracking technique that only estimates the three position DOFs (namely moving up and down, moving left and right, and moving forward and backward) is enough to provide a three-dimensional cursor-like input device driver \cite{Quesada2011a,Quesada2011b}. 

Such an input device could be used as a standard 2D mouse-like pointing device that considers depth changes to cause mouse-like clicks.
It also settles the bases for the development of virtual device drivers (i.e. software implemented device drivers, or not hardware device drivers) that consider three-dimensional position coordinates.

Real-time 3D motion tracking techniques have direct applications in several huge niche market areas \cite{Yilmaz2006}: the surveillance industry, which benefits from motion detection and tracking \cite{Kettnaker1999,Collins2001,Greiffenhagen2001}; the leisure industry, which benefits from novel human-computer interaction techniques \cite{Gallo2011}; the medical and military industries, which benefit from perceptual interfaces \cite{Bradski2000}, augmented reality \cite{Ferrari2001}, and object detection and tracking \cite{Forman2011}; and the automotive industry, which benefits from driver assistance systems \cite{Handmann1998}.

A 3D motion tracking system that only requires a single low-budget camera can be implemented in a wide spectrum of computers and smartphones that already have such a capture device installed.

There exist unsupervised markerless 3D motion tracking techniques \cite{Quesada2011a,Quesada2011b} that need no training, calibration, nor knowledge on the target object, and only require a single low-budget camera and an evenly colored object that is distinguishable from its surroundings.

These motion tracking techniques consist of a subsystem that determines the most relevant object in the screen, and a subsystem that performs the tracking by continuously estimating the target object projection features (center and area) from the edge image and a point inner to the object projection.

Existing object projection feature estimation techniques perform ray-casting from the inner point and estimate the center as the average of the ray hit location positions and the area as the coverage of the rays. Several variations of these techniques allow adjusting the number of rays being casted, the number of ray-casting iterations, and rasterization options. 
 
These techniques present three main drawbacks: when the inner point is surrounded by edges, rays may not reach other relevant areas of the object projection, which causes miscalculations in the object projection center and area estimations; the estimated features may greatly vary depending on the position of the inner point relative to the object projection; and finally, increasing the number of rays being casted and the ray-casting iterations to make the results more accurate and stable increases the processing time to the point the tracking cannot be performed on the fly.

In this paper, we analyze an intuitive filling-based object projection feature estimation technique that solves these issues and we study its main drawback: being too sensitive to edge miscalculations.
Then, we propose a less computing-intensive variant of that technique that makes it as less sensitive to edge miscalculations as ray-casting-based techniques are.

Section \ref{sec:back} covers existing object projection feature estimations techniques.
Section \ref{sec:fillgrid} describes and analyzes our two new approaches for solving the object projection feature estimation problem.
Finally, Section \ref{sec:concfw} summarizes the obtained conclusions and discloses the future work that derives from our research.

\section{Background} \label{sec:back}
\noindent

Unsupervised markerless 3D motion tracking techniques requires estimating the centroid and the area of the projection of a target object given an edge image and a point inside the object projection (namely, inner point) \cite{Quesada2011a,Quesada2011b}.
The inner point also has to be updated to increase the probabilities of it being inside the object projection in the next frame.
We call this the object projection feature estimation problem.

\begin{figure}[htb]
\centering
\includegraphics[scale=1]{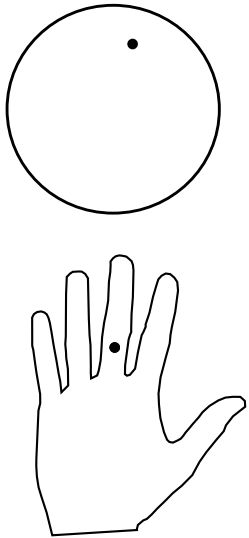}
\caption{The object projection feature estimation problem consists in, given an edge image and a point inside the object projection (namely, inner point), estimating the object projection centroid, the object projection area, and updating the inner point in order to increase the probabilities of it being inside the object projection in the next frame. Example of a convex object projection feature estimation problem (sphere projection) and to a non-convex object projection feature estimation problem (hand projection).}
\label{fig:feat1}
\end{figure}

Figure \ref{fig:feat1} depicts examples of a convex object projection feature estimation problem and a non-convex object projection feature estimation problem.

It should be noted that the inner point can be found enclosed in a small isolated area (e.g. a finger, when the target object is a hand).

It also should be noted that, due to the object movement between frames, it is possible for the current inner point to be relocated at a position that will be outside the object projection in the next frame.

Each one of the following subsections describes an approach for solving the object projection feature estimation problem.

\subsection{Feature Estimation Based on $n$-Ray-Casting} \label{sec:nray}
\noindent
Using this technique, $n$ rays are casted from the inner point position in different directions to hit an edge in the edge image \cite{Quesada2011a,Quesada2011b}.

The new centroid position is estimated to be the average of the ray hit location positions.

In order to estimate the inner point, it is displaced towards the new centroid until it reaches it or an edge. Then, rays are casted from the inner point and it is relocated at the average of the ray hit location positions, in order to center it in the projection area it is located, which reduces the probability of it being outside the object projection in the next frame.

The area is estimated to be the sum of the lengths of the casted rays.

\begin{figure}[htb]
\centering
\includegraphics[scale=1]{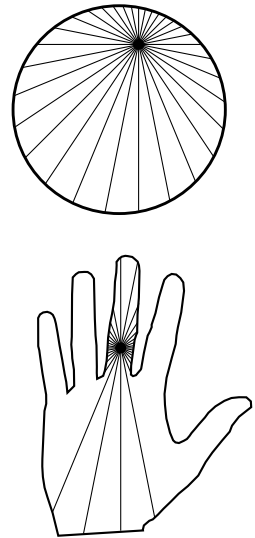}
\caption{$32$-ray-casting being applied to the estimation of the features of a convex object projection (sphere projection) and to a non-convex object projection (hand projection).}
\label{fig:feat2}
\end{figure}

Figure \ref{fig:feat2} illustrates $32$-ray-casting being applied to a convex object projection and to a non-convex object projection.

The main drawback of this technique is that the estimations may not be accurate when it is applied to non-convex object projections (e.g. a hand projection). In that case, the ray hit locations might be representative of just a fragment of the projection, in particular when the inner point is in a small isolated area of the object projection.
The centroid and the area might be inaccurately estimated, and the estimations may greatly vary depending on the position of the inner point relative to the object projection and on the ray orientations.

The likeliness of edge miscalculations (i.e. the edges not being calculated correctly) to have high impact in the projection area and centroid estimations is inversely proportional to $n$.

\subsection{Feature Estimation Based on Iterative $n$-Ray-Casting} \label{sec:inray}
\noindent
This technique is an iterative extension to $n$-ray-casting \cite{Quesada2011a,Quesada2011b}.

Using this technique, $n$ rays are casted from the inner point position in different directions to hit an edge in the edge image.

The new centroid position is estimated to be the average of the last iteration ray hit location positions.

The inner point is displaced towards the new centroid until it reaches it or an edge.

The process is repeated until the centroid and inner point adjustment is negligible or up to a maximum number of iterations.

Then, rays are casted from the inner point and it is relocated at the average of the ray hit location positions, in order to center it in the projection area it is located, which reduces the probability of it being outside the object projection in the next frame.

The area is estimated to be the sum of the rays casted during the last iteration.

\begin{figure}[htb]
\centering
\includegraphics[scale=1]{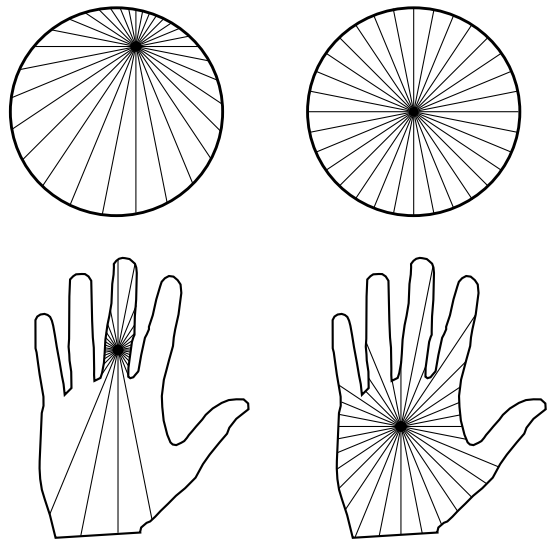}
\caption{Two steps of iterative $32$-ray-casting being applied to the estimation of the features of a convex object projection (sphere projection) and to a non-convex object projection (hand projection). Images on the left show the first iteration. Images on the right show the second iteration.}
\label{fig:feat2it}
\end{figure}

Figure \ref{fig:feat2it} illustrates two steps of iterative $32$-ray-casting being applied to a convex object projection and to a non-convex object projection.

It should be noted that iterative $n$-ray-casting can relocate the inner point into wider areas and therefore produce better estimations of the object projection centroid and area. Indeed, it can be observed that it produces better results than $n$-ray-casting when the target object is non-convex and the inner point is in a small isolated area of the target object projection.

Although this technique being iterative makes the centroid tend to be relocated into wider areas, the estimations are still not accurate when the technique is applied to non-convex object projections, as the ray hit locations might be representative of just a fragment of the object projection.

It should be noted that the centroid is not guaranteed to converge, and the estimations may greatly vary depending on the position of the inner point relative to the object projection, on the ray orientations, and on the maximum number of iterations.

The likeliness of edge miscalculations to have high impact in the projection area and centroid estimations is inversely proportional to $n$.

\subsection{Feature Estimation Based on Iterative $n^y$-Ray-Casting} \label{sec:inyray}
\noindent
This technique is an extension to iterative $n$-ray-casting \cite{Quesada2011a,Quesada2011b}.

Using this technique, $n$ rays are casted from the inner point position in different directions to hit an edge in the edge image.

Then, $n$ rays are casted from each of the last iteration ray hit location position. This re-casting process is repeated $y$ times for a total of $n^y$ rays being casted in the latest iteration.

The new centroid position is estimated to be the average of the last iteration ray hit location positions.

The inner point is displaced towards the new centroid until it reaches it or an edge.

Then, rays are casted from the inner point and it is relocated at the average of the ray hit location positions, in order to center it in the projection area it is located, which reduces the probability of it being outside the object projection in the next frame.

The process is repeated until the centroid and inner point adjustment is negligible or up to a maximum number of iterations.

The area is estimated to be the sum of the rays casted during the last iteration.

\begin{figure}[htb]
\centering
\includegraphics[scale=1]{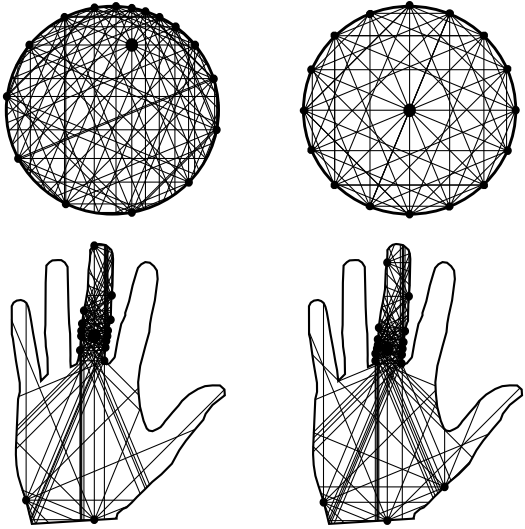}
\caption{Two steps of iterative $16^2$-ray-casting being applied to the estimation of the features of a convex object projection (sphere projection) and to a non-convex object projection (hand projection). Images on the left show the first iteration. Images on the right show the second iteration.}
\label{fig:feat3} 
\end{figure}

Figure \ref{fig:feat3} illustrates $16^2$-ray-casting being applied to a convex object projection and to a non-convex object projection.

It should be noted that the inner point is relocated into wider areas in non-convex object projections very slowly, due to isolated areas near the current inner point having a higher ray-density than wider areas, rendering the later less relevant for the estimation of the projection centroid and area.
On the other hand, iterative $n^y$-ray-casting covers the projection better than iterative $n$-ray-casting, and therefore outperforms it.

It should be noted that this technique, as $n$-ray-casting, does not guarantee the centroid to converge, and results may still greatly vary depending on the position of the inner point relative to the object projection, on the ray orientations, and on the maximum number of iterations.

The likeliness of edge miscalculations to have high impact in the projection area and centroid estimations is inversely proportional $n^y$. It should be noted that edge miscalculations near the inner point may produce very inaccurate results.

\subsection{Feature Estimation Based on Iterative $n^y$-Ray-Casting with $m$-Rasterization} \label{sec:inyrayr}
\noindent
This technique is an extension to iterative $n^y$-ray-casting \cite{Quesada2011a,Quesada2011b} that solves its problems.

Using this technique, $n$ rays are casted from the inner point position in different directions to hit an edge in the edge image.

Then, $n$ rays are casted from each of the last iteration ray hit location position. This re-casting process is repeated $y$ times for a total of $n^y$ rays being casted in the latest iteration.

Now, a rasterization process takes place. Every $m$x$m$ block that was run through by any of the rays is selected.

The new centroid position is estimated to be the average of the selected block positions.

The inner point is displaced towards the new centroid until it reaches it or an edge.

Then, rays are casted from the inner point and it is relocated at the average of the ray hit location positions, in order to center it in the projection area it is located, which reduces the probability of it being outside the object projection in the next frame.

The process is repeated until the centroid and inner point adjustment is negligible or up to a maximum number of iterations.
It should be noted that, as blocks always represent areas inside the object projection, no blocks are unselected between iterations.

The area is estimated to be the sum of the selected block areas.

\begin{figure}[htb]
\centering
\includegraphics[scale=1]{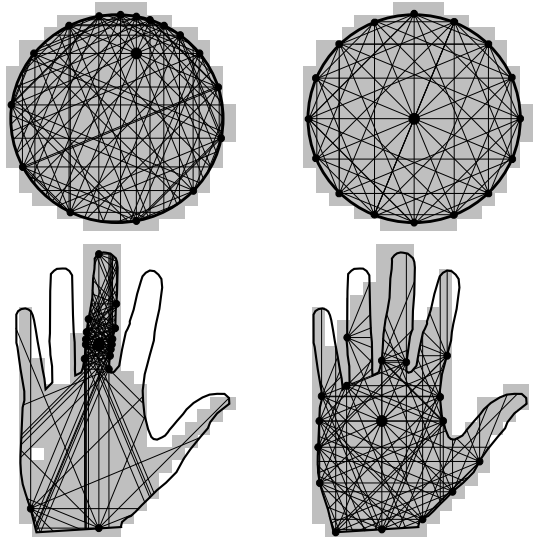}
\caption{Two steps of iterative $16^2$-ray-casting with $8$-rasterization being applied to the estimation of the features of a convex object projection (sphere projection) and to a non-convex object projection (hand projection). Images on the left show the first iteration. Images on the right show the second iteration.}
\label{fig:feat5}
\end{figure}

Figure \ref{fig:feat5} illustrates $16^2$-ray-casting with $8$-rasterization being applied to a convex object projection and to a non-convex object projection.

It should be noted that the inner point moves to wider areas in non-convex object projections quicker than when applying iterative $n^y$-ray-casting, due to high-ray-density areas being given the same relevance as low-ray-density areas. Less iterations are necessary for the estimations to be accurate, therefore processing times are lower than those of iterative $n^y$-ray-casting without rasterization, although they may still be prohibitive for certain applications.
It also should be noted that when $m$ is too high, the projection centroid and area estimations will be imprecise due to low resolution in block selection; when $m$ is too low, the technique behaves as iterative $16^2$-ray-casting without rasterization, which makes the inner point to be slowly displaced .

As the selected blocks are kept between iterations, the inner point and the centroid are guaranteed to converge. Although results may vary depending on the position of the inner point relative to the object projection, on the ray orientations, and on the maximum number of iterations, they will be similar for convex object projections and non-convex object projections with not too large isolated areas.

The likeliness of edge miscalculations to have high impact in the projection area and centroid estimations is inversely proportional to $n^y\cdot i$, being $i$ the number of performed iterations, as the final estimations depends on rays casted during any iteration.

\section{Filling-Based Techniques for Object Projection Feature Estimation} \label{sec:fillgrid}
\noindent
In this section, we propose the usage of filling-based techniques for solving the object projection feature estimation problem.

Filling-based techniques consist in locating every piece of the object projection until reaching the edges, and therefore fit the object projection better than ray-casting-based techniques.
Depending on the resolution and the parameters, filling-based techniques may also require a lower processing time than some of the more complex iterative ray-casting-based techniques.
It should be noted that, when using filling-based techniques, the feature estimations calculated from different inner points located in different parts of the same object will be the same.

In subsection \ref{sec:fill}, we analyze a simple filling-based object projection feature estimation technique.
In subsection \ref{sec:grid}, we propose an grid-based object projection feature estimation technique that outperforms all ray-casting-based techniques and the simple filling technique.

\subsection{Feature Estimation Based on Pixel-Filling} \label{sec:fill}
\noindent
This technique is an innovative approach to the object projection feature estimation problem that requires a lower processing time than ray-casting-based techniques.

Using this technique, the object projection is covered by filling it up to the edges.

A pixel queue is initialized with the inner point position pixel.
While the queue contains pixels, a pixel if extracted from the queue.
If the pixel is an edge in the edge image, it is ignored.
If the pixel is not an edge, it is marked as part of the object projection and all the pixels next to it that have not been marked are added to the queue.

The new centroid position is estimated to be the average of the marked pixel positions.

The inner point is displaced towards the new centroid until it reaches it or an edge.
Then, rays are casted from the inner point and it is relocated at the average of the ray hit locations, in order to center it in the projection area it is located, which reduces tracking errors.

The area is estimated to be the number of the marked pixels.

\begin{figure}[htb]
\centering
\includegraphics[scale=1]{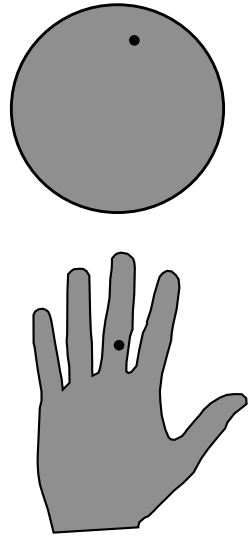}
\caption{Filling being applied to the estimation of the features of a convex object projection (sphere projection) and to a non-convex object projection (hand projection).}
\label{fig:feat8}
\end{figure}

Figure \ref{fig:feat8} illustrates filling being applied to a convex and a non-convex object projection.

It should be noted that this technique does not need to be iteratively applied, and the obtained results are the same independently of the inner point position.

However, this technique presents a major drawback that renders it unusable: edge miscalculations have high impact in the projection area and centroid estimations, as a single-pixel edge miscalculation would allow the filling to expand out of the actual object projection.

\subsection{Feature Estimation Based on $m$-Grid-Filling} \label{sec:grid}
\noindent
In order to solve the aforementioned pixel-filling technique problem, we propose a grid-filling technique.

Using this technique, a grid consisting of $m$x$m$ cells is casted centered in the inner point position and expanded until it reaches the edges.

A pixel queue is initialized with the inner point position pixel.
While the queue contains pixels, a pixel if extracted from the queue.
If the pixel is an edge in the edge image, it is ignored.
If the pixel is not an edge, it is marked as part of the object projection and all the pixels next to it that would be in the grid and that have not been marked are added to the queue.

The new centroid position is estimated to be the average of the grid pixel positions.

The inner point is displaced towards the new centroid until it reaches it or an edge.
Then, rays are casted from the inner point and it is relocated at the average of the ray hit locations, in order to center it in the projection area it is located, which reduces tracking errors.

The area is estimated to be the number of the grid pixel positions.

\begin{figure}[htb]
\centering
\includegraphics[scale=1]{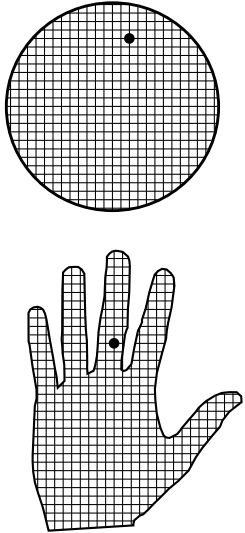}
\caption{$8$-grid-filling being applied to the estimation of the features of a convex object projection (sphere projection) and to a non-convex object projection (hand projection).}
\label{fig:feat7}
\end{figure}

Figure \ref{fig:feat7} illustrates grid-filling being applied to a convex and a non-convex object projection.

It should be noted that this approach solves all the problems of the aforementioned techniques: it is not as processing-time intensive as iterative $n$-ray-casting; it produces similar results independently of the inner point position; it allows the estimation of features of non-convex objects; it makes non-convex zones to be as relevant as convex-zones, as the grid pixel density is the same in the whole object projection; and the likeliness of edge miscalculations to have high impact in the projection area and centroid estimations is not as high as filling-based techniques'.

Also, as the grid size can be configured, the processing time requirements can be adjusted for low budget processor devices such as smartphones.
\section{Conclusions and Future Work} \label{sec:concfw}

In this paper, we have studied the object projection feature estimation problem in 3D motion tracking and we have studied the existing ray-casting-based techniques for solving it.

We have proposed an intuitive filling-based object projection feature estimation technique that produces a better coverage than ray-casting-based techniques. However, we have determined that this technique is not applicable in the practice since it is too sensitive to edge miscalculations.

Then, we have proposed a less computing-intensive grid-filling-based modification to that technique that solves the ray-casting-based techniques issues and is as sensitive to edge miscalculations as those techniques. Therefore, our technique can effectively replace ray-casting-based techniques.

Our proposal allows the development of more robust and less computing-intensive unsupervised markerless 3D motion tracking systems.

We plan to research on techniques that reduce the impact of edge miscalculations in feature estimation.

\bibliographystyle{plain}
\bibliography{doc}

\end{document}